\documentclass{article}

\usepackage{authblk}
\usepackage{natbib}
\setcitestyle{citesep={;}}

\usepackage[letterpaper,top=2cm,bottom=2cm,left=3cm,right=3cm,marginparwidth=1.75cm]{geometry}

\usepackage{amsmath}
\usepackage{graphicx}
\usepackage{xcolor}
\usepackage{url}
\usepackage{hyperref}
\hypersetup{
    colorlinks=true,
    linkcolor=purple,
    citecolor=blue,
    filecolor=magenta,      
    urlcolor=cyan,
}

\usepackage{lscape}

\usepackage[boxruled,vlined,linesnumbered]{algorithm2e}

\SetCommentSty{mycommfont}

\usepackage{amsmath}
\usepackage{amssymb}

\usepackage{booktabs}
\usepackage{multirow}

\let\emptyset\varnothing

\title{Structural Consequences of Policy-Based Interventions on the Global Supply Chain Network}

\author[1]{Lea Karbevska\thanks{Corresponding author. Email: lk547@cam.ac.uk}}
\author[1]{Liming Xu}
\author[1]{Zehui Dai}
\author[1]{Sara AlMahri}
\author[1,2]{Alexandra Brintrup}

\affil[1]{Department of Engineering, University of Cambridge, Cambridge, United Kingdom}
\affil[2]{The Alan Turing Institute, London, United Kingdom}

\date{}
\begin{document}
\maketitle

\begin{abstract}
As global political tensions rise and the anticipation of additional tariffs from the United States on international trade increases, the issues of economic independence and supply chain resilience become more prominent. 
The importance of supply chain resilience has been further underscored by disruptions caused by the COVID-19 pandemic and the ongoing war in Ukraine. 
In light of these challenges, ranging from geopolitical instability to product supply uncertainties, governments are increasingly focused on adopting new trade policies. This study explores the impact of several of these policies on the global electric vehicle (EV) supply chain network, with a particular focus on their effects on country clusters and the broader structure of international trade. Specifically, we analyse three key policies: Country Plus One, Friendshoring, and Reshoring. 
Our findings show that Friendshoring, contrary to expectations, leads to greater globalisation by increasing the number of supply links across friendly countries, potentially raising transaction costs. 
The Country Plus One policy similarly enhances network density through redundant links, while the Reshoring policy creates challenges in the EV sector due to the high number of irreplaceable products. 
Additionally, the effects of these policies vary across industries; for instance, mining goods being less affected in Country Plus One than the Friendshoring policy.
\end{abstract}

\section{Introduction}
\label{sec:introduction}
In recent years, supply chain disruptions have become increasingly frequent and severe due to multiple, compounding factors such as geopolitical tensions, the COVID-19 pandemic, tariffs, and climate-related events. 
This has intensified the need for robust and resilient supply networks capable of adapting to and recovering from such disruptions.\\

For example, the COVID-19 pandemic exposed significant weakness in global manufacturing and distribution networks \citep{covid_19}, leading to widespread shortages of personal protective equipment (PPE).
The ongoing Russia-Ukraine war has further disrupted global supply chains, prompting European countries to diversify their energy sources in an effort to reduce dependency on Russian energy imports \citep{energy_transition}. 
More recently, trade policies introduced during the Trump administration, such as tariffs on neighbouring countries and export bans on Chinese goods, are reshaping trade flows and underscoring the risks of overdependence on a single source for essential components.\\

These disruptions have promoted policymakers to consider interventions aimed at reshaping the structure of their national supply networks. 
For example, in the US, strategies such as Country Plus One (Country+1) \citep{china_plus_one, china_plus_one_strategy}, Reshoring \citep{reshoring_europe}, and the enforcement of new tariffs have been widely discussed.
Similarly, in the UK, approaches like Friendshoring and Nearshoring have been proposed as mitigation strategies \citep{near_sourcing,reshoring_uk}. 
While these policies may intuitively enhance supply chain robustness, they are fundamentally localized strategies, and their broader implications for global trade dynamics remain {\it uncertain}.\\

The analysis of supply chain network robustness has been an active topic of research \citep{IVANOV2024103081,AIBigDataSCR}. Prior studies have primarily examined the impacts of policies from focal firm perspectives, often focusing on potential cost savings or economic implications. For instance, in the case of Nearshoring \citep{reconsidering_nearshoring_2022} and Reshoring \citep{amico2024quantitative}, studies suggest these strategies significantly reduce transport costs. However, while financial analyses provide valuable insights, they often fail to capture the structural and systemic consequences these policies may have on the global supply chain network. The effects on network topology, robustness, and the propagation of disruptions through interconnected nodes have not been systematically analyzed, leaving a critical gap in understanding the broader ramifications of these interventions.\\

Over the past decades, emergent structural network properties have been identified as critical to the robustness of complex supply chains. One such property is the presence of scale-free or heavy-tailed structures in firm-to-firm connections, which make networks more robust to random failures but highly vulnerable to targeted failures at central hubs \citep{bacilieri2023firm, brintrup2016topological}. Similarly, nested structures within firm-to-firm and firm-to-product connections shelter unique products and links within large hubs, potentially enhancing robustness against localized failures \citep{brintrup2015nested}. Another key feature is modularity, which allows supply chains to form distinct communities, thereby containing disruptions within specific clusters or regions \citep{kito2014structure}.\\

Moreover, the distribution of domestic and international connections has a significant impact on network resilience. On average, supply networks exhibit a strong presence of domestic connections (30) compared to international ones (3), reducing the exposure of local production systems to global disruptions \citep{pichler2023building}. The negative assortativity observed in supply chain networks, where high-degree nodes tend to connect with low-degree nodes, helps distribute risk more evenly across the network and mitigates the likelihood of cascading failures \citep{bacilieri2023firm}. Additionally, the small-world properties of supply chain networks, characterized by higher clustering coefficients and shorter path lengths than expected in random networks, enhance robustness to random failures but simultaneously increase vulnerability to targeted attacks on hubs \citep{brintrup2015supply, galaskiewicz2011studying}.\\

Finally, the distributions of centrality measures, such as degree, betweenness, and closeness centrality, provide valuable insights into the roles of individual nodes in maintaining network connectivity and resilience. Nodes with high betweenness centrality, for instance, often serve as critical bridges within the network, and their removal can have disproportionately severe impacts on the overall performance of the supply chain \citep{kim2011structural}.\\

Understanding these systemic structural properties is crucial for designing global strategies aimed at enhancing the robustness and resilience of supply chain networks amid increasing uncertainties.\\

In this paper, we aim to bridge the gap between network analysis and policy consequences by exploring how localized supply network design interventions might reshape the broader network structure. 
Specifically, we investigate whether these interventions---designed to enhance national or regional resilience---could inadvertently introduce systemic inefficiencies or vulnerabilities. 
For example, policies encouraging multi-sourcing from allied or proximate countries may reduce reliance on a single supplier or region, but they could also create new clusters of interdependence that amplify risks within specific sectors. Similarly, policies promoting friendshoring might strengthen regional self-sufficiency but could also lead to fragmentation and a loss of economies of scale. 
Through this research, we seek to address these critical questions and provide a {\it first-of-its-kind} analysis of how such policies influence the structural robustness and resilience of global supply chains.\\

\begin{figure}[ht]
    \centering
    \includegraphics[width=1.1\textwidth]{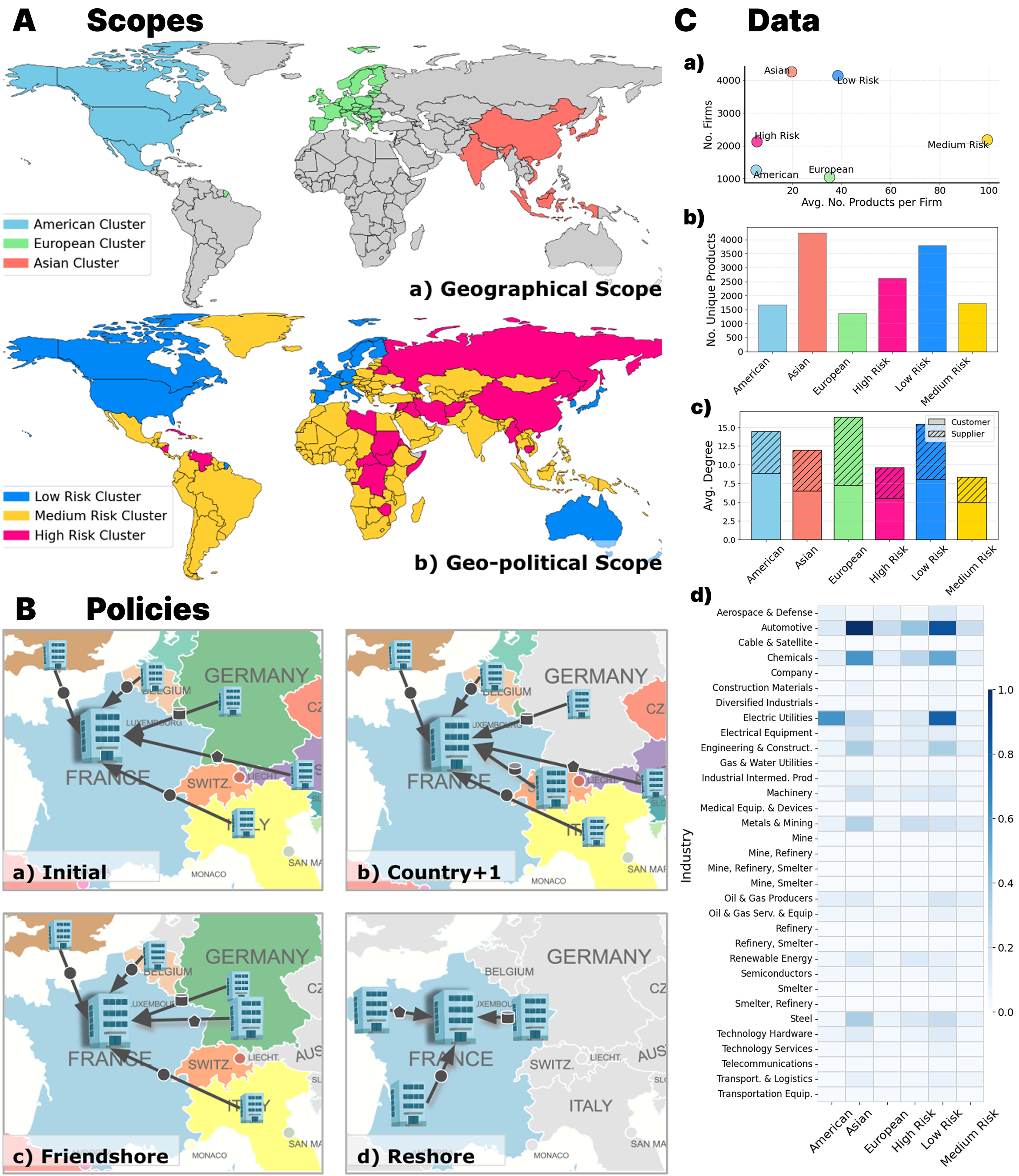}
    \caption{A) Illustration of the two groups of clusters used in the analysis: 
        (a) geographical clusters composed of a subset of American, Asian and European countries and
        (b) geopolitical clusters composed of  Low Risk cluster and Low \& Medium Risk cluster countries. 
    B) Examples illustrating the three policies: 
        (a) an initial supplier network to a company in France, 
        (b) Country+1 where the Country=Germany therefore France supplies from Germany and supplies the same products from a company outside of Germany, 
        (c) application of the policy Friendshoring where the company from France only supplies from neighbouring countries, and 
        (d) Reshore policy where the company only supplies from companies within the country. 
    C) Dataset statistics: 
        (a) average number of firms and products per firm, 
        (b) number of unique products, and 
        (c) in- and out-degrees, and specialisation of each cluster.
    }
    \label{fig:goal_data}
\end{figure}

We examine these policy impacts using supply chain data from the global EV industry, evaluating how the Country+1, Reshoring and Friendshoring policies, as illustrated in \autoref{fig:goal_data} B, reshape network structures. 
Specifically, we compute key network metrics and assess robustness across two dimensions (\autoref{fig:goal_data} A): 
{\it Geopolitical}, using the US three-tier risk system \citep{noauthor_biden_2025} for categorizing chip import tariffs, and {\it Geographical}, grouping major trade countries across three continents. 
Additionally, we identify vulnerable countries and products most affected by these interventions.\\


The rest of this paper is organized as follows. 
\autoref{sec:datasets} presents the datasets used in our analysis.
\autoref{sec:methodology} details our methodology, including the network metrics and the three policies. 
The analysis results are presented in \autoref{sec:results}. 
Finally, \autoref{sec:conclusions} discusses the results, concludes this paper and presents future work.

\section{Datasets}
\label{sec:datasets}
The data used for our analysis comes from three data sources: Bloomberg, Marklines (a private automotive industry database), and Mining Sites \citep{Jasansky2023}. 
All three data sources provide information on company-to-company supply relationships, including the specific products traded in Marklines and Mining Sites, and the industries covered in Bloomberg. 
Additionally, these sources offer details on company locations, market capitalisations, and the size of the relationships between companies. The Bloomberg data, in particular, was collected between July and December 2023, ensuring the analysis is based on recent and relevant information.\\

We focus on the EV industry because of its complexity, global scale, and the significant reliance of auto manufacturers on overseas suppliers for a diverse range of inputs, including raw materials from mining. The combined dataset comprises 18,833 companies, 66,745 different product types (e.g. Lithium, Engine Temperature Sensor, and Software) sourced from across 32 industries (e.g. Automotive, Aerospace \& Defense, Chemicals, Mine, Refinery, and Smelter) and 149 countries (China, Canada, Spain, and Russia, etc.). For more details on the dataset, refer to \autoref{subsec:appendix_dataset}.\\

However, a significant portion of the customer-supplier relationships in our dataset lacked detailed product information. To address this gap, we used a pre-trained large language model (LLM) to extract product data from company websites, as described in \citep{almahri_enhancing_2024}, thereby augmenting our dataset. \\

We then compiled a list of product category names, totalling 1,102 categories. This list was constructed by combining Marklines product names (929 categories), Bloomberg industry classifications outside the automotive sector (32 categories), and products from the Mining Sites dataset (141 categories).\\

To map the 66,745 product names in our dataset to these categories, we leveraged three pretrained LLMs: 
{\tt bert-base-nli-mean-tokens}, 
{\tt paraphrase-MiniLM-L6-v2}, and 
{\tt roberta-large}. 
The mapping was determined by either the majority agreement across the three models or, in cases without a majority, the category with the highest similarity score provided by any of the models. 
This approach ensured robust and consistent categorisation of product names.

\section{Methodology}
\label{sec:methodology}
In this study, we conceptualize supply chain data as a network, where nodes represent firms and directed edges capture the supplier-customer relationships between them. Each node is further characterized by the products it manufactures.\\

We compute a subset of structural metrics known to influence supply chain resilience, such as degree distribution, nestedness, modularity, assortativity, small-world properties, and centrality measures, to evaluate the network's vulnerability and robustness under different configurations \citep{brintrup2015nested, brintrup2016topological, kito2014structure, bacilieri2023firm, galaskiewicz2011studying, kim2011structural, pichler2023building}.\


In particular, we analyse these metrics of the supply chain network before and after the application of the three policies: Country+1, Friendshoring, and Reshoring. These policies impose constraints on the geographical origins of suppliers of companies within a specific country or a group of countries. To model them, we apply two distinct country clustering strategies.\\

The first clustering strategy, \textit{Geographical},  divides major trade countries into three continental clusters: 
North American, European, and Asian. 
The North American cluster includes the US, Mexico, and Canada; the European cluster includes countries such as those in the European Union, Switzerland, and the UK; and the Asian cluster encompasses nations like China, India, Japan, Indonesia, and others. 
These two classification strategies are illustrated in \autoref{fig:goal_data} A (a) and \autoref{fig:goal_data} A (b) respectively.\\

The second clustering strategy, \textit{Geopolitical}, is based on the three-tier risk system established by the US to categorize countries for tariff purposes concerning chip imports. This system classifies countries into three risk levels: Low Risk includes close allies such as South Korea and Taiwan, which have unrestricted access to advanced AI chips; Medium Risk includes countries like those in Eastern Europe and Latin America, which face restrictions on-chip access; and High Risk includes nations such as China and Russia, which are entirely banned from receiving US chips \citep{noauthor_biden_2025}. This classification is depicted in \autoref{fig:goal_data} A (a).\\

In addition, during each simulation, we track the number of companies and products for which companies within a given cluster are unable to find alternative suppliers whilst adhering to the geographical constraints imposed by the specific policy. 
Given the stochastic nature of our methodology, each simulation is conducted over five iterations, and the average values are reported to ensure the robustness and reliability of the results.\\

It is important to emphasize that real world conditions are likely to reflect a mixture of these scenarios. 
The results presented here correspond to extreme rewiring cases derived from the applied policies.

\subsection{The Three Policies}
This section details the three policies examined in our analysis: Country Plus One (Country+1), Friendshoring, and Reshoring. The detailed pseudo-code for each policy algorithm is provided in \autoref{subsec:policy_pseudo}.

\subsubsection{Country Plus One}\label{sec:country_1}
The Country+1 policy, as illustrated in \autoref{fig:goal_data} B(b), involves designating a specific country or a list of countries, \textit{X}, and seeking additional suppliers outside of \textit{X} for products currently sourced from there, thereby reducing dependence on country \textit{X} \citep{china_plus_one, china_plus_one_strategy}.\\

For this particular scenario, motivated by recent geopolitical developments in the US and the broader Western context \citep{noauthor_biden_2025,standard_western_2025,jackson_china_2025}, we designate \textit{X} as China in the \textit{geographical} scope. This decision reflects the significant reliance on China for critical components in the automotive industry, which is the focus of our dataset.\\

Additionally, within our \textit{geopolitical} scope, we designate \textit{X}  as the High Risk countries (such as China, Russia, Iran, and North Korea) to assess the extent of global dependence on these nations. This approach aligns with the US government's efforts to address economic and national security risks associated with reliance on certain countries for critical goods and materials \citep{usa_gov_critical_mineral}.\\

\autoref{fig:goal_data} B b) illustrates the Country+1 policy, where, for illustrative purposes, the scope is on a company in France, and Germany is selected as Country \textit{X}.

\subsubsection{Friendshoring}
Friendshoring or Nearshoring (see \autoref{fig:goal_data} B (c)) is a policy designed to reallocate suppliers to geographically closer or geopolitically aligned countries, enhancing supply chain stability.\\

Nearshoring involves sourcing products and services from suppliers in geographically proximate regions \citep{near_sourcing}. Its primary economic benefits stem from shorter transport corridors, leading to reduced logistics costs and lower emissions. Additionally, shorter delivery times enhance flexibility, allowing businesses to respond more efficiently to fluctuations in demand. This strategy serves as a middle ground between full globalization and economic isolation, as well as between offshoring and domestic production (reshoring).\\

Friendshoring, on the other hand, prioritizes sourcing from suppliers located in politically allied or strategically aligned countries \citep{hayashi2022nations}. Rather than solely focusing on economic efficiency, this approach emphasizes long-term stability and supply chain resilience by reducing exposure to geopolitical risks.\\

In our scenarios, we define ``friends'' as countries belonging to the same cluster (e.g., Low Risk, Asian and European). In the geographical scope, these clusters generally comprise geographically proximate countries. In contrast, under the geopolitical scope, countries within the same cluster are not necessarily geographically close.\\

Refer to \autoref{fig:goal_data}B(c) for a visual illustration of the Friendshoring policy, focusing on a company based in France. 
In this example, friendshoring entails relocating suppliers to neighboring countries.

\subsubsection{Reshoring}
The Reshoring policy, as illustrated in \autoref{fig:goal_data} B (d), aims to bring production and outsourcing services back to the home country \citep{reshoring_uk,reshoring_europe}. This policy was especially discussed during the COVID-19 pandemic since many countries could not rely on other countries and international logistics and started diversifying their economy and bringing back at home production. In this analysis, we explore the extent to which each country can feasibly reshore its outsourced goods and services. \\

Refer to \autoref{fig:goal_data} B (d) for a visual example of the Reshoring policy, where, for illustrative purposes, the scope is on a company in France, and the restoring policy process involves relocating suppliers to France.\\

\section{Results}
To evaluate the impact of the proposed policies on the EV supply chain network, we compute key topological metrics before and after their implementation. We begin by examining the initial structural characteristics of the real-world supply chain network, followed by an analysis of the changes induced by the simulated policy interventions.

\label{sec:results}

\subsection{Initial Network Properties}
\label{subsec:initial_results}
The initial EV supply chain network, prior to policy implementation, exhibits several key characteristics. 
The network has a heavy-tailed structure in the firm-to-firm connections, which helps mitigate random failures. The nodes on average have 1.87 international and 1.86 domestic connections, which means firms rely on very few suppliers and customers, potentially making the network vulnerable. Additionally, the relatively balanced international-to-domestic connections indicate higher exposure to global disruptions in the automotive sector compared to other industries \citep{acemoglu2012network}.\\

The network's modularity of 0.57 indicates a strong community structure, which may help contain disruptions within clusters and prevent spreading globally. Like other supply chain networks, the network exhibits negative degree assortativity (-0.12), meaning high-degree firms (hubs) tend to connect with low-degree firms. 
However, we also observe positive location assortativity, suggesting firms prefer local connections. \\

The clustering coefficient is 0.026, implying that firms are not strongly interconnected within local groups. The average shortest path length of 4.57 suggests that disruptions can propagate relatively quickly across the network, but not as fast as in highly interconnected small-world networks.\\

The most central companies based on eigenvector and closeness centrality include Toyota, Volkswagen, Honda, Stellantis, and Mercedes-Benz, highlighting their dominant role in the network. These firms act as critical bridges and influencers in the supply chain, meaning disruptions at these nodes could have widespread effects.\\

\begin{figure}[ht]
\centering
    \includegraphics[width=1\textwidth]{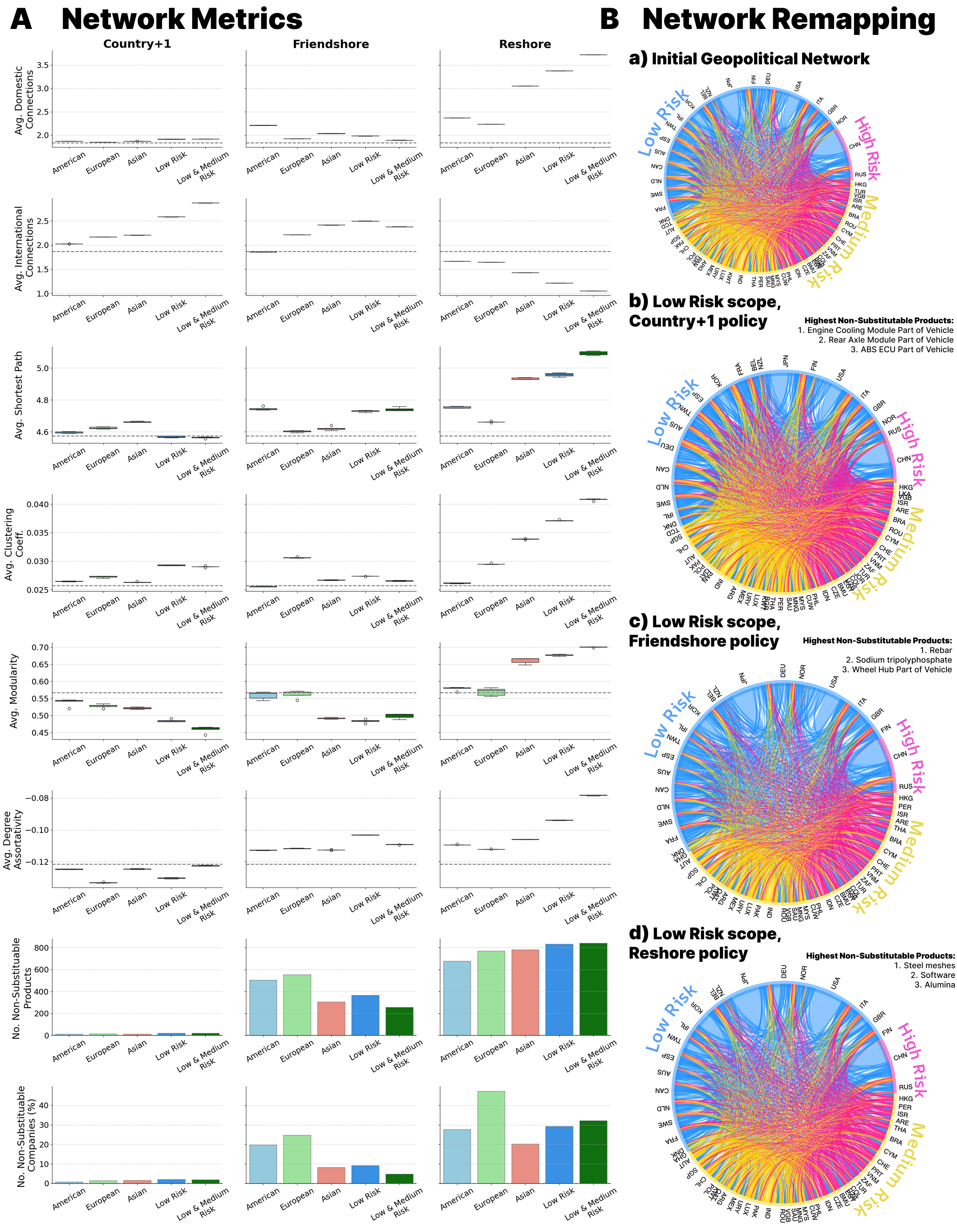}
    \caption{
        A) Visualization of results across five network metrics: shortest path, domestic and international connections, modularity, size of non-substitutable products and companies' suppliers across five clusters (American, European, Asian, Low Risk, and Low \& Medium Risk). 
        The results are displayed as box plots for indeterministic metrics (metrics that change due to the randomness of our policy applications) and as bar plots for those with deterministic behaviour. 
        The grey dotted line represents the parameter value in the initial network before any policy application. 
        B) Illustration of the initial mapping (a) and the remapping (b, c, d) after the application of each policy on the Low Risk scope between firms of different countries grouped by the geopolitical clusters.
    }
\label{fig:results}
\end{figure}

\begin{figure}[ht]
\centering
    \includegraphics[width=1\textwidth]{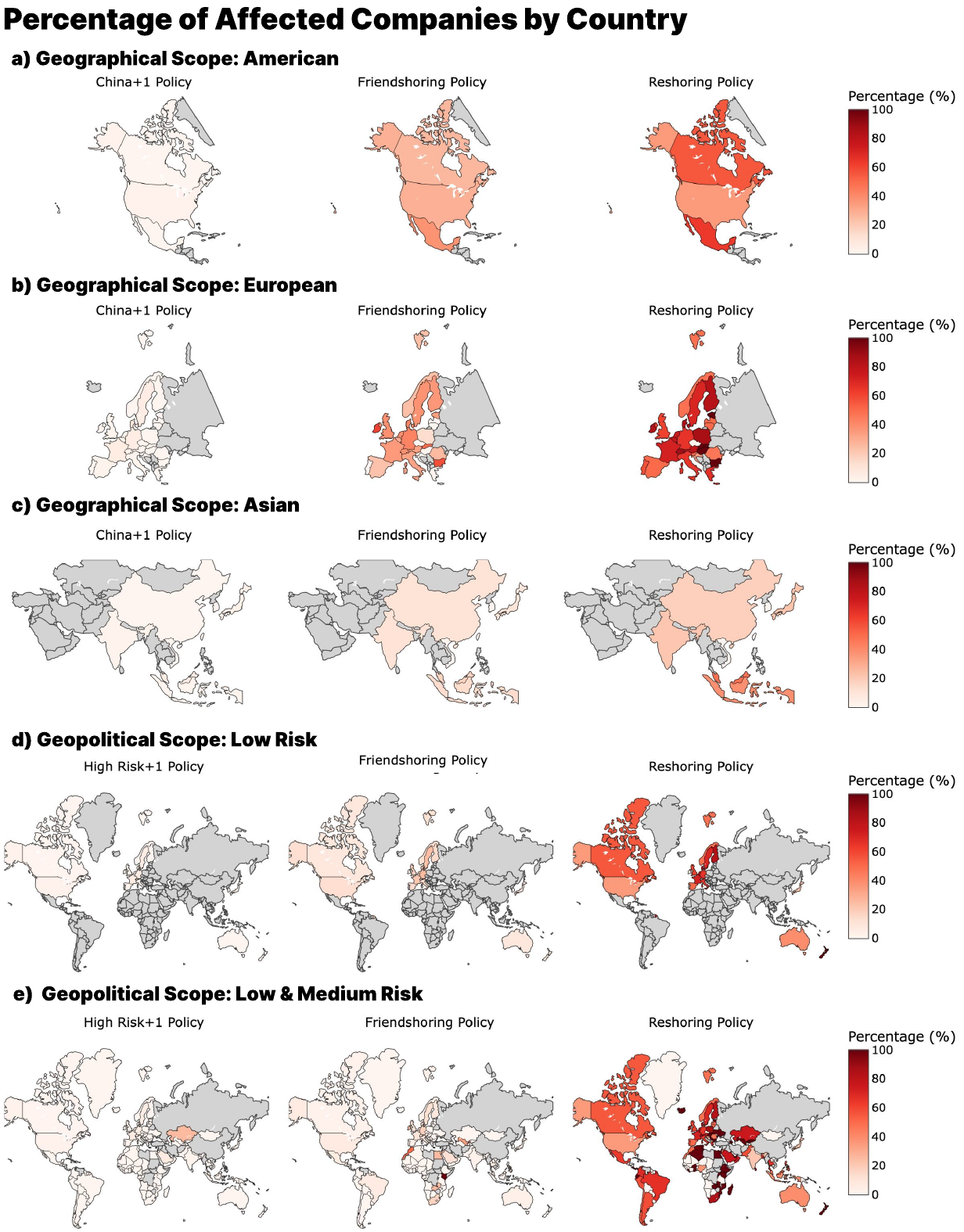}
    \caption{
        Visualization of the number of affected companies by country after the application of the policies Country+1 (China+1 and High Risk+1), Friendshoring and Reshoring across the Geographical scope (American, European, Asian) and Geopolitical scope (Low Risk, and Low \& Medium Risk).
    }
\label{fig:results_country}
\end{figure}

\subsection{Policy Simulations}
Next, we analyze network changes under different policy applications. In the Geographical scope, we assess the effects of China+1 (Country+1), Friendshoring, and Reshoring on the American, European, and Asian clusters. In the Geopolitical scope, we examine the impact of High Risk+1 (Country+1), Friendshoring, and Reshoring on the Low Risk and Low \& Medium Risk clusters.\\

We present our findings through multiple perspectives: changes in overall supply chain network metrics (\autoref{fig:results}), the policy impact on each country by number of affected companies (\autoref{fig:results_country}), as well as the top five most vulnerable industries (\autoref{Tab:VulnerableIndustries}) across scopes and policies. Further results examining product- and country-level vulnerabilities are presented in \autoref{subsec:appendix_results}.

\subsubsection{Geographical Scope}
To compute the results in the geographical scope, we focused on three distinct clusters: the \textit{American}, \textit{European} and \textit{Asian}. 
For a detailed list of the countries included in these clusters and our dataset, see~\autoref{subsec:scopes}.\\

For each cluster, we apply three trade policies: China+1 (Country+1), Friendshoring, and Reshoring.\\
First, we apply the \textit{Country+1} policy, where ``Country'' refers specifically to \textit{China}. This policy is designed to reduce dependency on Chinese suppliers by incorporating additional suppliers from other countries.\\

Under the China+1 policy, the \textit{American} cluster contains 13 non-substitutable products—items that cannot be sourced from any company outside China (\autoref{fig:results}A). These products are primarily concentrated in the Transportation Equipment, Industrial Intermediates, and Semiconductors sectors (\autoref{Tab:VulnerableIndustries}). Notably, 21 American companies rely on these non-substitutable products, making the US the only country with firms affected under the China+1 policy in the American cluster (\autoref{fig:results_country}a).\\

The \textit{European} cluster shows 14 non-substitutable products, a number that remains unchanged after excluding mining products (\autoref{fig:results}A). The top three industries contributing to these non-substitutable products are Transportation Equipment, Automotive, and Technology Services (\autoref{Tab:VulnerableIndustries}). The countries with the highest number of companies with products in the non-substitutable list are Germany (16 companies), France (6 companies), and Spain (3 companies) (\autoref{fig:results_country}b).\\

In the \textit{Asian} cluster, 12 non-substitutable products are identified (\autoref{fig:results}A). The most affected industries are Transportation Equipment, Telecommunications, and Construction Materials (\autoref{Tab:VulnerableIndustries}). The countries with the highest number of companies with non-substitutable products are Taiwan (65 companies), Japan (25 companies), and South Korea (10 companies) (\autoref{fig:results_country}c).\\

In the China+1 policy, we observe that the American cluster is the least dependent on China, while the Asian cluster, particularly Hong Kong and Taiwan, is the most dependent. Across all three clusters, there is a noticeable increase in international connections, a reduction in modularity, and an increase in the shortest path, particularly in the Asian cluster.\\

Next, when applying the Friendshoring policy on the American cluster, we observe 503 non-substitutable products, where 54 are mining products (\autoref{fig:results}A). The most impacted industries are Transportation Equipment, Oil \& Gas Services \& Equipment, and Industrial Intermediate Products(\autoref{Tab:VulnerableIndustries}). The United States has the highest number of affected companies, 369 in total, followed by Canada (70 companies) and then Mexico (21 companies) (\autoref{fig:results_country}a).\\

The European cluster reveals 552 non-substitutable products, with 58 mining products (\autoref{fig:results}A). The most impacted industries are Cable \& Satellite, Telecommunications, and Metals \& Mining (\autoref{Tab:VulnerableIndustries}). The countries with the highest number of companies affected are Germany (123 companies), the United Kingdom (99 companies), and Spain (51 companies) (\autoref{fig:results_country}b).\\

The Asian cluster shows 307 non-substitutable products, with 26 mining products (\autoref{fig:results}A). Key industries affected are Cable \& Satellite, Telecommunications, and Industrial Intermediate Products (\autoref{Tab:VulnerableIndustries}). China (224 companies), Japan (123 companies), and South Korea (81 companies) are the countries with the highest number of affected companies in the Asian cluster (\autoref{fig:results_country}c).\\

In the Friendshoring policy, we notice that the Asian cluster exhibits the most resilience among the three clusters and has the highest increase in international links and the highest decrease in modularity, while the European cluster is the most vulnerable, with the highest number of companies and products being affected. The American Cluster, on the other hand, has a decrease in international connections and the highest increase in domestic connections.\\

Lastly, when applying the Reshoring policy on the American cluster contains 676 non-substitutable products, with 119 mining products (\autoref{fig:results}A). Mining, Refining, and Smelting industries contribute the most to the non-substitutable products (\autoref{Tab:VulnerableIndustries}). The United States has the highest number of affected companies (460 companies), then Canada (148 companies) and Mexico (36 companies) (\autoref{fig:results_country}a).\\

The European cluster has 769 non-substitutable products, with 133 mining products (\autoref{fig:results}A). The most affected industries are Mining, Refining, and Smelting (\autoref{Tab:VulnerableIndustries}). Countries with the highest number of affected companies are Germany (192 companies), the United Kingdom (158 companies), and Spain (113 companies) (\autoref{fig:results_country}b).\\

In the Asian cluster, there are 780 non-substitutable products, with 136 mining products (\autoref{fig:results}A). The Mining, Refining, and Smelting industries show the highest proportion of non-substitutable products (\autoref{Tab:VulnerableIndustries}). The countries with the most company suppliers affected are China (378 companies), Japan (287 companies) and Taiwan (237 companies) (\autoref{fig:results_country}c).\\

In the Reshoring policy, we notice an increase in domestic connections, an increase in modularity, and a decrease in international connections, which is expected. We notice that the countries in the European cluster are the most vulnerable, while the Asian cluster has substantially the least affected percentage of companies, yet we acknowledge that those companies act as bottlenecks which receive supplies of various products.\\

When comparing the resilience of the different regional clusters to the China+1, Friendshoring, and Reshoring policies, we observe distinct patterns of impact. Our findings reveal that the European cluster demonstrates the least resilience to Friendshoring and Reshoring policies, with substantial disruption in key industries. In contrast, the American cluster shows the highest resilience to the China+1 policy, with fewer products in the non-substitutable list across the board. The Asian cluster, while being the most dependent on China, is the most resilient to Friendshoring with the lowest percentage of companies and number of products affected. In the case of Reshoring, while we notice that the Asian cluster has the lowest percentage of companies affected, we still acknowledge that those companies act as bottlenecks which receive supplies of various products.

\begin{table}[h]
\centering
\caption{Top 5 most vulnerable industries under each policy scenario across different scopes.}
\label{Tab:VulnerableIndustries}
\resizebox{\textwidth}{!}{%
\begin{tabular}{lllll}
\toprule
\multicolumn{1}{c}{\textbf{Scope / Policy}} & \textbf{China+1} & \textbf{Friendshoring} & \textbf{Reshoring} \\ \midrule

\multirow{5}{*}{American} 
& Transportation Equipment      & Transportation Equipment     & Mine, Refinery, Smelter \\
& Industrial Intermediate Prod  & Oil \& Gas Services \& Equip & Mine, Smelter \\
& Semiconductors                & Industrial Intermediate Prod & Mine \\
& Technology Services           & Steel                        & Smelter \\
& Renewable Energy              & Machinery                    & Telecommunications \\ \midrule

\multirow{5}{*}{European} 
& Transportation Equipment      & Cable \& Satellite           & Mine, Refinery, Smelter \\
& Automotive                   & Telecommunications           & Mine, Smelter \\
& Technology Services           & Metals \& Mining             & Refinery, Smelter \\
& Construction Materials        & Industrial Intermediate Prod & Smelter \\
& Industrial Intermediate Prod  & Machinery                    & Cable \& Satellite \\ \midrule

\multirow{5}{*}{Asian} 
& Transportation Equipment      & Cable \& Satellite           & Mine, Refinery, Smelter \\
& Telecommunications            & Telecommunications           & Mine, Smelter \\
& Construction Materials        & Industrial Intermediate Prod & Refinery, Smelter \\
& Aerospace \& Defense          & Transportation \& Logistics  & Smelter \\
& Automotive                   & Metals \& Mining             & Smelter, Refinery \\ \midrule

\multicolumn{1}{c}{\textbf{Scope / Policy}} & \textbf{High Risk+1} & \textbf{Friendshoring} & \textbf{Reshoring} \\ \midrule

\multirow{5}{*}{Low Risk} 
& Transportation Equipment      & Industrial Intermediate Prod & Mine, Refinery, Smelter \\
& Automotive                   & Automotive                   & Mine, Smelter \\
& Industrial Intermediate Prod  & Machinery                    & Refinery, Smelter \\
& Construction Materials        & Transportation Equipment     & Smelter \\ 
& Renewable Energy              & Electrical Equipment         & Cable \& Satellite \\ \midrule

\multirow{5}{*}{Low \& Medium Risk} 
& Transportation Equipment      & Machinery                    & Mine, Refinery, Smelter \\
& Automotive                   & Automotive                   & Mine, Smelter \\
& Construction Materials        & Telecommunications           & Refinery, Smelter \\
& Technology Services           & Industrial Intermediate Prod & Smelter \\
& Semiconductors                & Diversified Industrials      & Cable \& Satellite \\

\bottomrule
\end{tabular}%
}
\end{table}

\subsubsection{Geopolitical Scope}
To compute the results within the geopolitical scope, we focused on two distinct clusters: \textit{Low Risk} and \textit{Low \& Medium Risk}. The latter aggregates countries from both the Low Risk and Medium Risk groups to provide a more comprehensive global perspective. For a detailed list of the countries included in the two aforementioned clusters and our dataset, see \autoref{subsec:scopes}. For each of these clusters, we apply three trade policies: High Risk+1 (Country+1), Friendshoring, and Reshoring. These policies are analyzed to evaluate their respective impacts on the overall network and the clusters of supply chain networks.\\

First, we apply the Country+1 policy, where ``Country" refers to all companies within the High Risk cluster. The policy aims to reduce reliance on 
high risk suppliers by incorporating additional suppliers from outside this group. In this case, both\textit{ Low Risk} and Low \& Medium Risk exhibit a similar number of non-substitutable products, 21 and 22, respectively (\autoref{fig:results}A).\\

The top industries contributing to these non-substitutable products in both cases are Transportation Equipment and Automotive, while Mining, Refining, Smelting, Diversified Industrials, and Cable \& Satellite industries remain unaffected (\autoref{Tab:VulnerableIndustries}). Additionally, the countries with the highest number of companies in the non-substitutable list are Taiwan (65 companies), Japan (26 companies), and the United States (21 companies) (\autoref{fig:results_country}d,e). \\

While the policy increases global connections and reduces modularity in both clusters, applying the \textit{High Risk+1} policy to the Low \& Medium Risk cluster results in the highest increase in international links and the lowest modularity among all cases (\autoref{fig:results}). This outcome aligns with expectations, as adding more suppliers enhances network robustness. However, companies in Taiwan appear to be the most vulnerable, indicating that while the policy diversifies the supplier base, certain countries, like Taiwan, remain highly dependent on suppliers from High Risk cluster.\\

Next, when applying the Friendshoring policy, the Low Risk cluster reveals 367 non-substitutable products, with 9.54\% belonging to the mining sector (\autoref{fig:results}A). The most impacted industries are Industrial Intermediate Products, Automotive, and Machinery (\autoref{Tab:VulnerableIndustries}). The countries with the highest number of affected companies are the United States (170 companies), Japan (118 companies), South Korea (73 companies), and Germany (73 companies) (\autoref{fig:results_country}d).\\

In the Low \& Medium Risk cluster, the number of non-substitutable products decreases to 256, with mining products accounting for 8.59\% (\autoref{fig:results}A). The most impacted industries are Machinery, Automotive, and Telecommunications, while Mining, Refining, and Smelting remain unaffected (\autoref{Tab:VulnerableIndustries}). The highest number of affected suppliers have the companies in Japan (77 companies), the US (75 companies) and Germany (52 companies) (\autoref{fig:results_country}e).\\

Notably, expanding from Low Risk to Low \& Medium Risk significantly reduces the number of affected products and companies, mitigating the vulnerability of the United States and Japan by broadening their supplier base. Interestingly, although this policy restructures supplier relationships within a predefined cluster of countries, it still promotes internationalization as evidenced by the significant increase in international links, while decreasing modularity in the supply network.\\

Lastly, when applying the Reshoring policy, the Low Risk cluster contains 831 non-substitutable products, with 137 mining products (\autoref{fig:results}A). The most affected industries are Mining, Refining, and Smelting, and Cable \& Satellite (\autoref{Tab:VulnerableIndustries}). The countries with the highest number of companies with non-substitutable products are the United States (460 companies), Japan (287 companies), and Taiwan (237 companies) (\autoref{fig:results_country}d). \\

In the Low \& Medium Risk cluster, the number of non-substitutable products slightly increases to 840, with mining products accounting for 16.31\% (\autoref{fig:results}A). The most affected industries remain Mining, Refining, Smelting (\autoref{Tab:VulnerableIndustries}). The United States, Japan, and Taiwan continue to be the countries with the highest number of affected companies (\autoref{fig:results_country}e). \\

Extending the application of the Reshoring policy from Low Risk to Low \& Medium Risk increases the number of affected products and companies, though the most vulnerable countries remain unchanged. Notably, this policy reduces international connections while increasing domestic ones. It also results in the highest network modularity and the longest average shortest path, reinforcing company clustering. Additionally, we observe that mining products and raw materials emerge as the most vulnerable category under this policy, highlighting the risks associated with localizing supply chains for critical resources.









\section{Discussion and Conclusions}\label{sec:conclusions}
In this paper, we seek to bridge the gap between network analysis and the consequences of policy interventions by examining how localized supply network design changes impact broader network structures. We aim to identify patterns that emerge at a systemic level, shedding light on how these policies reshape global supply chains.\\

Existing studies that explore the systemic implications of localized, geopolitically motivated supply chain policies are relatively scarce, and those that do exist often lack firm- and product-level data. To address this gap, we collected large-scale data on critical minerals in the electric vehicle sector and tested the extremes of three key policies: Country+1, Friendshoring, and Reshoring.\\

We find that, contrary to common intuition, Friendshoring leads to greater globalization. As firms replace supplier links with companies outside their scope that have a wide product range, countries pursuing the policy form more links across a network of ``friendly'' nations to replace all the products, leading to a higher density of connections. This may lead to consequences such as increased transaction costs, as firms need to manage additional supply chain relationships.\\

The Country+1 policy, which aims to introduce redundancy into supply chains, also results in increased connectivity and higher network density. 
However, as expected, it does not generate a substantial number of non-substitutable products. In contrast, an extreme implementation of the Reshoring policy appears infeasible in the current EV battery sector, given the prevalence of irreplaceable products—particularly in upstream segments such as mining. \\

Moreover, there is a significant heterogeneity in how these policies affect different industries, influenced by the tiered structure of the global supply network. For instance, mining products are less affected by the Country+1 policy, as they tend to be preserved, whereas they are significantly impacted by Reshoring.\\

In conclusion, our study highlights the need for greater nuance in understanding the effects of these policies. Product-level data is crucial to determine how systemic structures evolve at the global scale. Given the variability in product spaces and firm-level connections across industries, we believe our findings have broader implications. Future research should expand this analysis beyond the electric vehicle sector and examine how regionalization impacts the robustness of global supply chains. Furthermore, studies exploring the changes in firms' transaction costs due to policy shifts and analyzing the trade-offs between regionalization and supply chain resilience are necessary.\\

While our analysis is focused on a subset of supply chains within the EV sector and a specific time period, this work serves as an initial attempt to quantify and provide insights into the effects of the three supply chain intervention policies: Country+1, Friendshoring, and Reshoring. The variations in resilience observed across different regional clusters underscore the complexity of these policies and highlight the need for a deeper understanding of their impact on global supply chains. We hope that this research lays the groundwork for future studies that expand the analysis to other supply chains and more comprehensive datasets, ultimately offering a broader perspective on how such policies reshape global supply network dynamics.

\subsection{Data Availability}
The data used in this analysis includes proprietary datasets from Bloomberg and Marklines. While we are unable to share these datasets directly due to licensing restrictions, they are commercially available and can be obtained through the respective providers. Additionally, we use the Mining Sites dataset, which is publicly available as part of the study by \citet{Jasansky2023}.

\subsection{Code Availability}
We provide pseudocode (\autoref{subsec:policy_pseudo}) to illustrate the policy simulation framework. The code implementation can be shared upon reasonable request.

\clearpage


\bibliographystyle{plainnat}
\bibliography{references}

\section{Acknowledgments}
\textbf{Contribution:} We thank Dimitris Skarlatoudis, Bang Xiang Yong, Alex Clinton, and Yunbo Long for assistance with data collection and validation. 
We also thank the members of the Supply Chain AI Lab and the participants of the 5th Interdisciplinary Workshop on Firm-Level Supply Networks in Oxford for their valuable input and comments.

\section{Author Contributions}
\textbf{L.K.}: Led the writing, implemented and simulated the policies, and created the visualisations. 
\textbf{L.X.}: Contributed to the data cleaning and the writing. 
\textbf{Z.D.}: Developed the initial code for simulating policy applications and contributed to the data cleaning. 
\textbf{S.A.}: Completed the missing product data using her LLM-based software. \textbf{A.B.}: Contributed to refining the writing and guided research ideas and overall direction.

\newpage
\appendix
\setcounter{page}{1}         
\renewcommand{\thepage}{A\arabic{page}} 

\section{Appendix}
\label{sec:appendix}

\subsection{Dataset}
\label{subsec:dataset}
\label{subsec:appendix_dataset}
\begin{figure}[ht]
    \centering
    \includegraphics[width=1\textwidth]{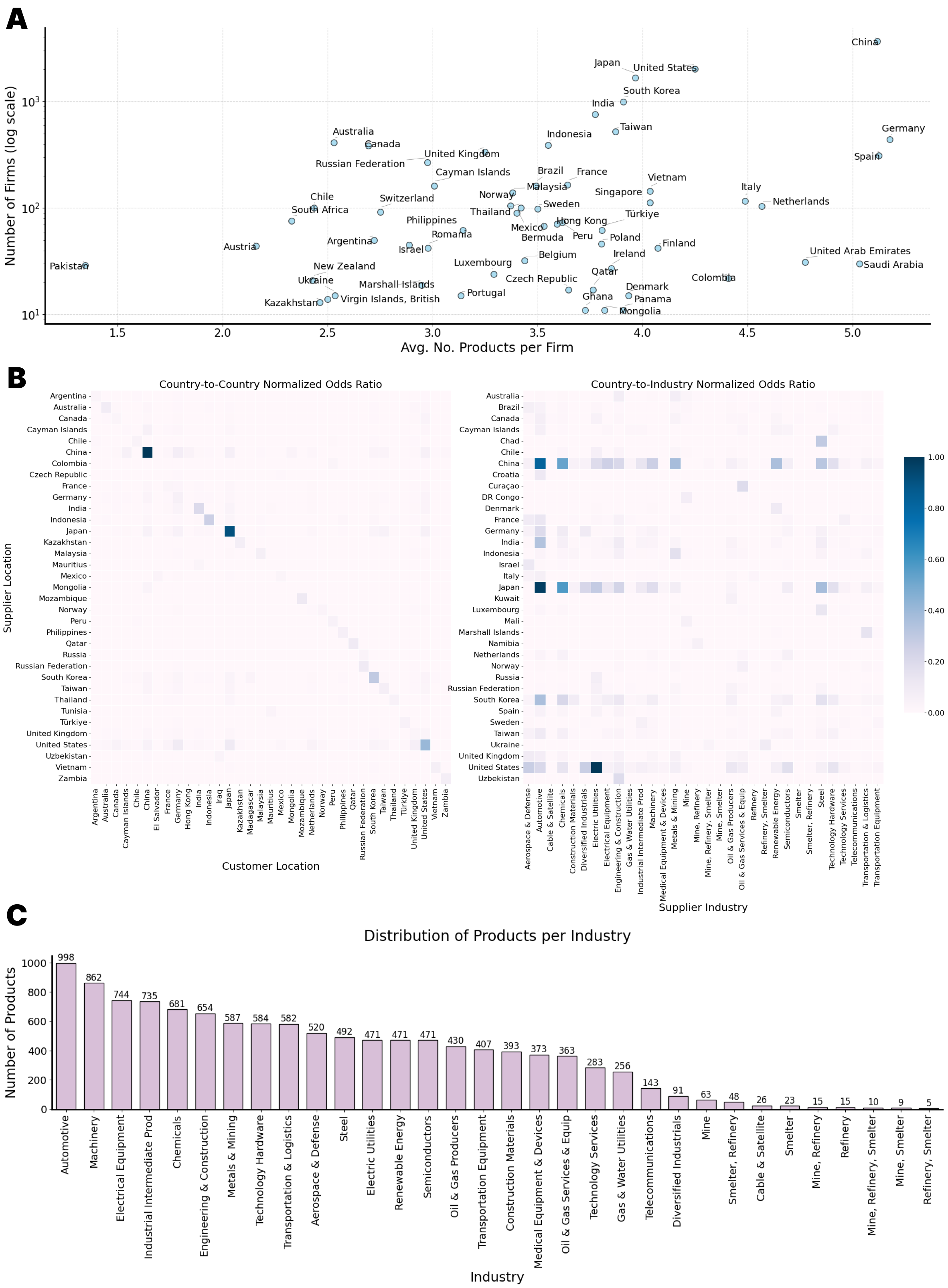}
    \caption{
        Visualisation of our dataset.  
        \textbf{(a)} Scatter plot showing the number of firms ($x$-axis) and the average number of products per firm ($y$-axis) for all locations with more than 10 firms.
        \textbf{(b)} Heatmaps: the left panel illustrates bilateral customer-supplier relationships between countries, and the right panel depicts country-industry specialisation in production.
        \textbf{(c)} Distribution of the number of products across industries.
    }
\label{fig:data}
\end{figure}

In our dataset, as shown in \autoref{fig:data}(a), China has the highest number of companies (i.e., 2111 companies), followed by Japan (1325) and the US (1322). 
On average, companies in this dataset manufacture around 5 products. 
In terms of supplier-customer relationships, trade tends to occur domestically within countries.
Regarding country-product specialisation, companies from the US predominantly supply electric utility products, while firms from Japan and China tend to focus on automotive products.
The distribution of the number of products across industries is presented in \autoref{fig:data}(c), with the Automotive and Machinery industries containing the largest number of products.

\subsection{Policy Pseudocode}\label{subsec:policy_pseudo}
To enhance clarity, this section details the specific algorithmic procedures employed to implement each of the four policies.

\subsubsection{Country+1 Policy}
\label{subsec:appendix_countryplusone}
The Country+1 policy, as presented in \autoref{alg:plusone}, takes two input arguments: sets of countries \( X \) and \( S \), where companies in \( S \) aim to reduce their dependence on suppliers from \( X \). 
The algorithm begins by extracting two sets of companies:
\( C_X \), comprising companies located in risky countries \( X \) (e.g., suppliers from China and Russia);
\( C_{S} \), representing companies operating in \( S \).
To monitor supply chain vulnerabilities, the algorithm utilises two sets: \( NS_p \), which stores non-substitutable products, and \( NS_c \), which contains companies that cannot find alternative suppliers within \( S \). \\

The main process involves iterating over each supplier \( x \) in \( C_X \) to check whether companies in \( C_S \) rely on it. 
For each company \( c \) in \( C_S \), the algorithm retrieves the products \( P \) supplied by \( x \). 
It then examines whether each product \( p \) in \( P \) can be sourced from an alternative supplier in friendly countries \( S \). 
If at least one alternative supplier exists, it is added to \( c \)’s supplier list (unless it already exists). 
If no suitable alternative supplier is found, the product is marked as non-substitutable and added to \( NS_p \) and the company is added to \( NS_c \), flagging its reliance on suppliers from \( X \).  
After enforcing this policy, companies in \( S \) will have alternative suppliers wherever possible, thus reducing their dependency on \( X \). 
As a byproduct, this algorithm would provide insights into supply chain risks by identifying non-substitutable products and companies that remain reliant on suppliers from risky countries, enabling the development of more targeted risk mitigation strategies.

\begin{algorithm}[t]
\small
\DontPrintSemicolon
\SetKwInOut{Input}{input} 
\SetKwInOut{Output}{output} 
\SetKwFunction{GetProducts}{GetProducts}
\SetKwFunction{GetSuppliers}{GetSuppliers}
\SetKwFunction{GetCompanies}{GetCompanies}
\caption{Procedure of Enforcing the Country+1 Policy}\label{alg:plusone}
\Input{ $X$ is the countries with risks, e.g., Russia and China; $S$ is friendly countries (excluding $X$) for seeking additional suppliers}
\Output{Remapped network after enforcing Country+1 policy}
$C_X \gets \GetCompanies{X}$ \;
$C_S \gets \GetCompanies{S}$ \;
$NS_p \gets \emptyset$ \tcp*{a set of non-substitutable products}
$NS_c \gets \emptyset$ \tcp*{a set of companies with non-substitutable suppliers}
\For{$x \in C_X$}{ 
    \For{$c \in C_S$}{ 
        \For{$x \in \GetSuppliers{c}$}{
            $P \gets \GetProducts{x}$ \tcp*{Products supplied by company $x$}
            \For{$p \in P$}{
                $alternativeSuppliers \gets \emptyset$ \;
                \For{$s \in C_S$}{
                    \If{$s \neq c \And p \in \GetProducts{s}$}{
                        \text{add} $c$ \text{to} $alternativeSuppliers$ \; 
                    }
                }
                \If{$alternativeSuppliers = \emptyset$}{
                    \text{add} $p$ \text{to} $NS_p$ \;
                    \text{add} $c$ \text{to} $NS_c$ \;
                }
                \Else{
                    \For{$s \in alternativeSuppliers$}{
                        \If{$s \notin \GetSuppliers{c}$}{
                            \text{add} $s$ \text{as a supplier of} $c$ \tcp*{add alternative supplier}
                        }
                    }
                }
            }   
        }
    }
}
\end{algorithm}

\subsubsection{Friendshoring Policy}
\label{app:friendshoring}

The Friendshoring (\autoref{alg:nearshore}) policy enforces companies in a given set of allied or neighbouring countries \( S \) to replace foreign suppliers with local alternatives wherever possible. 
The algorithm begins by retrieves all companies in \( S \), and initialises sets to track non-substitutable products and companies that cannot find their alternative suppliers. 
For each company \( c \) in \( S \), the algorithm examines its suppliers and identifies those located outside of \( S \). 
It then checks whether each product supplied by an external supplier has at least one alternative supplier within \( S \). 
If such a substitute exists, the company updates its supply chain to add this new local supplier; otherwise, the product and the associated company are flagged as non-substitutable. 
To ensure minimal disruption, an external supplier is only removed if all its products can be replaced.
The output of this algorithm is a refined supply chain where companies have increased reliance on local suppliers, while non-substitutable products and dependencies are tracked for further intervention.

\begin{algorithm}[th]
\small
\DontPrintSemicolon
\SetKwInOut{Input}{input} 
\SetKwInOut{Output}{output} 
\SetKwFunction{GetProducts}{GetProducts}
\SetKwFunction{GetSuppliers}{GetSuppliers}
\SetKwFunction{GetCompanies}{GetCompanies}
\caption{Procedure of Enforcing the Friendshoring Policy}\label{alg:nearshore}
    \Input{A set of countries that are allies or friends$S$}
    \Output{Remapped network after intervention}
    \BlankLine
    $C \gets \GetCompanies{S}$ \tcp*{Get all companies in countries $S$}
    $NS_p \gets \emptyset $ \tcp*{Non-substitutable products}
    $NS_c \gets \emptyset $ \tcp*{Companies with non-substitutable suppliers}

    \For{$c \in C$}{
        \For{$s \in \GetSuppliers{c}$}{
            \If{$s \notin C$}{
                $P \gets \GetProducts{s} $ \tcp*{Get products supplied by $s$ and assign them to $P$}
                $substitutable \gets \textbf{true}$ \;
                $alternativeSuppliers \gets \emptyset $ \;
                \For{$p \in P$}{ 
                    \tcp*[l]{Check if this product can be substituted; if so, find alternative suppliers}
                    \For{$ k \in C $}{
                        \If{$ k \neq c \And p \in \GetProducts{k} $}{
                        \text{add} $k$ \text{to} $alternativeSuppliers$ \;
                    }
                    }
                    \If{$alternativeSuppliers = \emptyset$  }{
                        \text{add} $p$ \text{to} $NS_p$ \;
                        \text{add} $c$ \text{to} $NS_c$ \;
                        $substitutable \gets \textbf{false}$ \;
                    }
                    \Else{
                        \For{$k \in alternativeSuppliers$}{
                            \tcp*[l]{Add $k$ as a supplier of $c$}
                            \If{$k \notin \GetSuppliers{c}$}{
                                \text{add} $k$ \text{as a supplier of} $c$\;
                            } 
                        }
                    }
                }
                \tcp*[l]{If substitution exists, remove this supplier $s$}
                \If{$ substitutable = \text{\bf true} $}{
                    \text{remove} $s$ \text{as a supplier of} $c$\;
                }
            }
        }
    }
\end{algorithm}

\subsubsection{Reshoring Policy}
\label{app:reshoring}

The Reshoring  (\autoref{alg:reshore}) aims to replace foreign suppliers with domestic alternatives for companies operating in a given set of countries \( S1 \). It first retrieves all companies in \( S1 \) and initializes sets to track non-substitutable products and companies that cannot replace their foreign suppliers. For each company \( c \) in \( S1 \), it identifies its home country and examines its current suppliers. If a supplier \( s \) is foreign (i.e., located outside \( c \)’s home country), the algorithm checks whether each product supplied by \( s \) has at least one alternative domestic supplier within \( S1 \). If a substitute exists, the company adds the new local supplier; otherwise, the product and company are marked as non-substitutable. Finally, a foreign supplier is only removed if all its products can be replaced, ensuring minimal disruption. The output provides a refined supply chain where companies depend more on local suppliers, while non-substitutable products and dependencies are tracked for further intervention.

\begin{algorithm}[t]
\small
\DontPrintSemicolon
\SetKwInOut{Input}{input}
\SetKwInOut{Output}{output}
\SetKwFunction{GetCompanies}{GetCompanies}
\SetKwFunction{GetSuppliers}{GetSuppliers}
\SetKwFunction{GetProducts}{GetProducts}
\SetKwFunction{GetCountry}{GetCountry}
\SetKwFunction{AddSupplier}{AddSupplier}
\SetKwFunction{RemoveSupplier}{RemoveSupplier}
\caption{Procedure of Enforcing the Reshoring Policy}\label{alg:reshore}
    \Input{A set of countries $S1$}
    \Output{Remapped network after reshoring intervention}
    \BlankLine
    $C \gets \GetCompanies{S1}$ \tcp*{Get all companies in $S1$}
    $NS_p \gets \emptyset$ \tcp*{Non-substitutable products}
    $NS_c \gets \emptyset$ \tcp*{Companies with non-substitutable suppliers}

    \For{$c \in C$}{
        $home \gets \GetCountry{c}$ \;
        \For{$s \in \GetSuppliers{c}$}{
            \If{$\GetCountry{s} \neq home$}{
                $P \gets \GetProducts{s}$ \;
                $substitutable \gets \textbf{true}$ \;
                \For{$p \in P$}{
                    $alternatives \gets \emptyset$ \;
                    \For{$k \in C$}{
                        \If{$k \neq c \And \GetCountry{k} = home \And p \in \GetProducts{k}$}{
                            \text{add} $k$ \text{to} $alternatives$ \;
                        }
                    }
                    \If{$alternatives = \emptyset$}{
                        \text{add} $p$ \text{to} $NS_p$ \;
                        \text{add} $c$ \text{to} $NS_c$ \;
                        $substitutable \gets \textbf{false}$ \;
                    }
                    \Else{
                        \For{$k \in alternatives$}{
                            \If{$k \notin \GetSuppliers{c}$}{
                                \AddSupplier{$c, k$} \;
                            }
                        }
                    }
                }
                \If{$substitutable = \textbf{true}$}{
                    \RemoveSupplier{$c, s$} \;
                }
            }
        }
    }
\end{algorithm}

\subsection{Scenario Scopes}
\autoref{Tab:scopes} presents the complete set of countries in the Asian, American, European, Low Risk, Medium Risk, and High-Risk clusters used in our experiments.
\label{subsec:scopes}
\begin{landscape}
\begin{table}[t]
\centering
\caption{List of countries categorized by their Geopolitical (Low, Medium, and High Risk) and Geographical (Asian, American, and European) clusters, which have at least one customer or supplier company in the dataset.}
\label{Tab:scopes}
\resizebox{0.925\columnwidth}{!}{%
\begin{tabular}{llllllll}
\toprule
\multicolumn{4}{c}{\textbf{Geopolitical Clusters}} &
   &
  \multicolumn{3}{c}{\textbf{Geographical Clusters}} \\ \cmidrule(lr){1-5}\cmidrule(lr){6-8}
\textbf{Low Risk (19)} &
  \multicolumn{3}{c}{\textbf{Medium Risk (107)}} &
  \textbf{High Risk (15)} &
  \textbf{Asian (10)} &
  \textbf{American (3)} &
  \textbf{European (31)} \\ \cmidrule(lr){1-1}\cmidrule(lr){2-4}\cmidrule(lr){5-5}\cmidrule(lr){6-6}\cmidrule(lr){7-7}\cmidrule(lr){8-8}
Australia      & Algeria            & Ghana            & Namibia              & Belarus            & China       & Canada        & Austria        \\
Belgium        & Angola             & Greece           & New Caledonia        & Cambodia           & Hong Kong   & Mexico        & Belgium        \\
Canada         & Argentina          & Guatemala        & Nigeria              & China              & India       & United States & Bulgaria       \\
Denmark        & Aruba              & Guinea           & Oman                 & DR Congo           & Indonesia   &               & Croatia        \\
Finland        & Austria            & Guyana           & Pakistan             & Cuba               & Japan       &               & Cyprus         \\
France         & Azerbaijan         & Honduras         & Panama               & Iran               & Malaysia    &               & Czech Republic \\
Germany        & Bahrain            & Hong Kong        & Papua New Guinea     & Iran               & Singapore   &               & Denmark        \\
Ireland        & Bangladesh         & Hungary          & Peru                 & Iraq               & South Korea &               & Estonia        \\
Italy          & Bermuda            & Iceland          & Philippines          & Lebanon            & Taiwan      &               & Finland        \\
Japan          & Bhutan             & India            & Poland               & Libya              & Vietnam     &               & France         \\
Netherlands    & Bolivia            & Indonesia        & Portugal             & Myanmar            &             &               & Germany        \\
New Zealand    & Bosnia-Herzegovina & Israel           & Puerto Rico          & North Korea        &             &               & Greece         \\
Norway         & Brazil             & Jamaica          & Qatar                & Russian Federation &             &               & Hungary        \\
South Korea    & Bulgaria           & Jordan           & Romania              & Venezuela          &             &               & Ireland        \\
Spain          & Burkina Faso       & Kazakhstan       & Saudi Arabia         & Zimbabwe           &             &               & Italy          \\
Sweden         & Cameroon           & Kenya            & Serbia               &                    &             &               & Latvia         \\
Taiwan         & Cayman Islands     & Kuwait           & Singapore            &                    &             &               & Lithuania      \\
United Kingdom & Chad               & Kyrgyzstan       & Slovakia             &                    &             &               & Luxembourg     \\
United States  & Chile              & Laos             & Slovenia             &                    &             &               & Malta          \\
               & Colombia           & Latvia           & South Africa         &                    &             &               & Netherlands    \\
               & Costa Rica         & Liberia          & Sri Lanka            &                    &             &               & Norway         \\
               & Croatia            & Liechtenstein    & Suriname             &                    &             &               & Poland         \\
               & Curaçao            & Lithuania        & Switzerland          &                    &             &               & Portugal       \\
               & Cyprus             & Luxembourg       & Tanzania             &                    &             &               & Romania        \\
               & Czech Republic     & Madagascar       & Thailand             &                    &             &               & Slovakia       \\
               & Czechia            & Malaysia         & Tunisia              &                    &             &               & Slovenia       \\
               & Côte d'Ivoire      & Mali             & Türkiye              &                    &             &               & Spain          \\
               & Dominica           & Malta            & Uganda               &                    &             &               & Sweden         \\
               & Dominican Republic & Marshall Islands & Ukraine              &                    &             &               & Switzerland    \\
               & Ecuador            & Mauritania       & United Arab Emirates &                    &             &               & United Kingdom \\
               & Egypt              & Mauritius        & Uruguay              &                    &             &               &                \\
               & El Salvador        & Mexico           & Uzbekistan           &                    &             &               &                \\
               & Estonia            & Moldova          & Vietnam              &                    &             &               &                \\
               & Ethiopia           & Mongolia         & Virgin Islands       &                    &             &               &                \\
               & Fiji               & Morocco          & Zambia               &                    &             &               &                \\
               & Ghana              & Mozambique       &                      &                    &             &               &                \\ \hline
\end{tabular}%
}
\end{table}
\end{landscape}

\subsection{Additional Results}
\label{subsec:appendix_results}
\begin{figure}[ht]
\centering
    \includegraphics[width=1\textwidth]{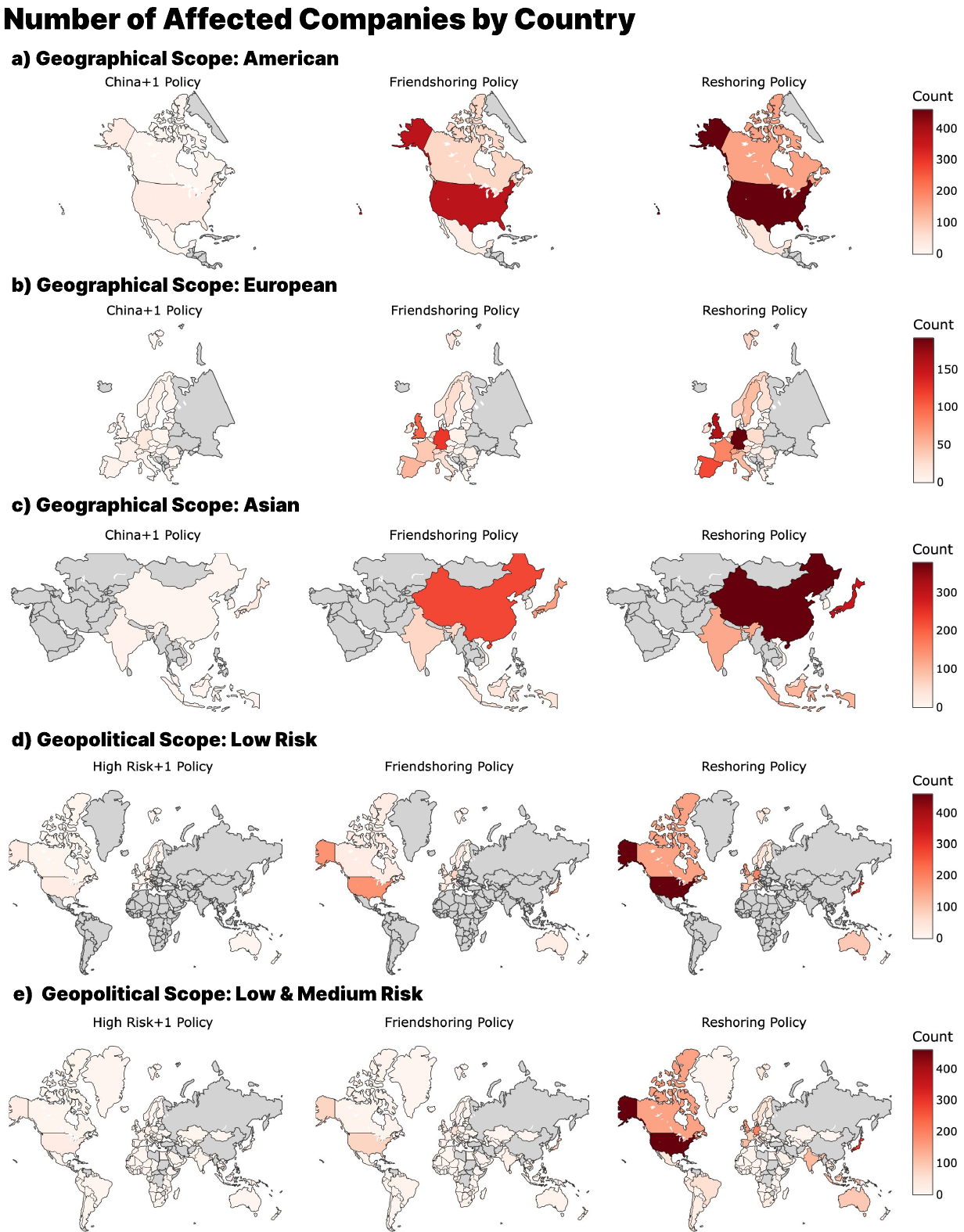}
    \caption{
        Visualization of the percentage of affected companies by country after the application of the policies Country+1 (China+1 and High Risk+1), Friendshoring and Reshoring across the Geographical scope (American, European, Asian) and Geopolitical scope (Low Risk, and Low \& Medium Risk).
    }
\label{fig:results_country_percentage}
\end{figure}

\begin{figure}[ht]
    \centering
    \includegraphics[width=0.9\textwidth]{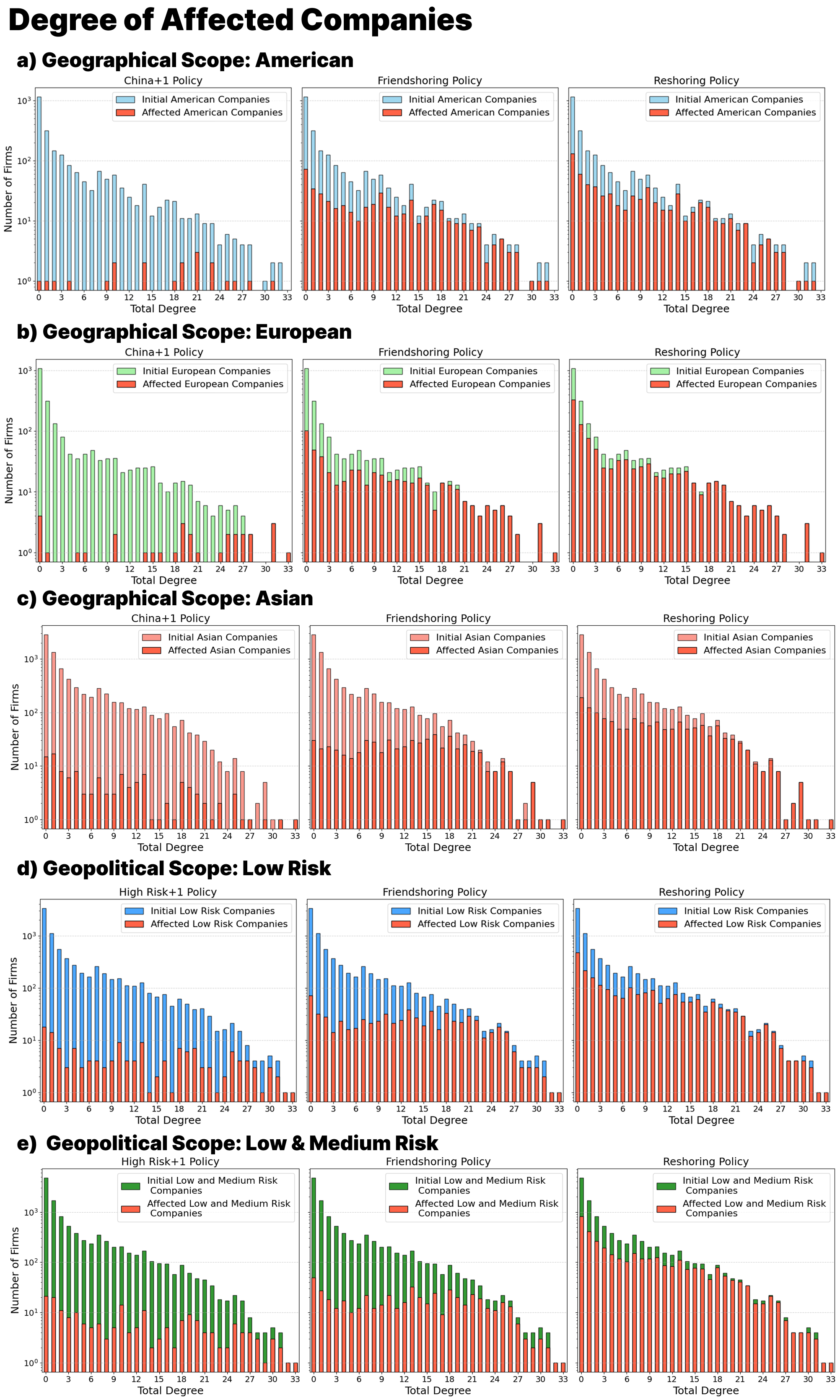}
    \caption{
    Visualization of the degree of the affected companies after the application of the
policies Country+1 (China+1 and High Risk+1), Friendshoring and Reshoring across the Geographical and Geopolitical scope}
\label{fig:results_company_degree}
\end{figure}

To complement the results discussed above, we report in \autoref{Tab:American}, \autoref{Tab:European}, and \autoref{Tab:Asian} the global network metrics for the Geographical Scope, and in \autoref{Tab:Tier1} and \autoref{Tab:Tier12} for the Geopolitical Scope, comparing the networks before and after policy-induced rewiring. These metrics provide a structural overview of how each policy reshapes the supply chain. Building on this, \autoref{Tab:VulnerableProducts} identifies the five most vulnerable products across all policy–scope combinations, highlighting where disruptions may concentrate. The broader vulnerability patterns at the country level are shown in \autoref{fig:results_country_percentage}, which plots the share of companies to find an alternative supplier under the respective policy constraints. Finally, \autoref{fig:results_company_degree} examines whether such vulnerabilities are more prevalent among highly connected hubs or smaller firms, by comparing the degree distribution of all companies to that of the affected subset.

\subsubsection{Geographical Scope}
\label{subsec:appendix_result_geographical}

\begin{table}[h]
\centering
\caption{Network measures before and after enforcing the three policies in the American cluster.}
\label{Tab:American}
\resizebox{\textwidth}{!}{%
\begin{tabular}{lllll}
\toprule
\multicolumn{2}{l}{\textbf{}}                                 & \multicolumn{3}{c}{\textbf{After enforcing policy}}              \\ \cmidrule{3-5} 
\textbf{Network Measures}                 & \textbf{Initial} & \textbf{China+1} & \textbf{Friendshoring} & \textbf{Reshoring} \\ \midrule
Avg. No. Edges                  & 56339     & 59287.2   & 61931     & 61478     \\
Density                         & 0.000243  & 0.000256  & 0.000268  & 0.000266  \\
Degree Assortativity            & -0.121478 & -0.124697 & -0.112675 & -0.109282 \\
Location Assortativity          & 0.428589  & 0.410186  & 0.47501   & 0.52449   \\
Avg. Shortest Path Length       & 4.575134  & 4.597169  & 4.745798  & 4.753068  \\
Avg. Domestic connections       & 1.836291  & 1.870452  & 2.212053  & 2.372831  \\
Avg. International Connections  & 1.866325  & 2.02592   & 1.85807   & 1.667521  \\
Clustering Coefficient          & 0.025749  & 0.02649   & 0.025562  & 0.026149  \\
Modularity                      & 0.565294  & 0.539261  & 0.559002  & 0.578745  \\
No. Communities                 & 1438      & 1269.6    & 1457.2    & 1450.4    \\
No. Non-Substitutable Products (Mines \%) & 0                 & 13 (0\%)         & 503 (10.74\%)           & 676 (17.6\%)        \\
No. Non-Substitutable Companies & 0         & 21        & 476       & 664       \\ \bottomrule
\end{tabular}%
}
\end{table}

\begin{table}[h]
\centering
\caption{Network measures before and after enforcing the three policies in the European cluster.}
\label{Tab:European}
\resizebox{\textwidth}{!}{%
\begin{tabular}{llccc}
\toprule
\multicolumn{2}{l}{\textbf{}}                                 & \multicolumn{3}{c}{\textbf{After enforcing policy}}              \\ \cmidrule{3-5} 
\textbf{Network Measures}                 & \textbf{Initial} & \textbf{China+1} & \textbf{Friendshoring} & \textbf{Reshoring} \\ \midrule
Avg. No. Edges                  & 56339     & 61148     & 62989     & 59018     \\
Density                         & 0.000243  & 0.000264  & 0.000272  & 0.000255  \\
Degree Assortativity            & -0.121478 & -0.133112 & -0.111633 & -0.112158 \\
Location Assortativity          & 0.428589  & 0.395352  & 0.409187  & 0.525181  \\
Avg. Shortest Path Length       & 4.575134  & 4.624493  & 4.602967  & 4.661914  \\
Avg. Domestic Connections       & 1.836291  & 1.849501  & 1.926354  & 2.235673  \\
Avg. International Connections  & 1.866325  & 2.169164  & 2.213302  & 1.643007  \\
Clustering Coefficient          & 0.025749  & 0.027267  & 0.030633  & 0.029528  \\
Modularity                      & 0.56117   & 0.527753  & 0.562455  & 0.569449  \\
No. Communities                 & 1440      & 1205.4    & 1472.2    & 1445.2    \\
No. Non-Substitutable Products (Mines \%) & 0         & 14 (0\%)   & 552 (10.51\%) & 769 (17.3\%) \\
No. Non-Substitutable Companies & 0         & 32        & 519       & 993       \\ \bottomrule
\end{tabular}%
}
\end{table}

\begin{table}[h]
\centering
\caption{Network measures before and after enforcing the three policies in the Asian cluster.}
\label{Tab:Asian}
\resizebox{\textwidth}{!}{%
\begin{tabular}{lllll}
\toprule
\multicolumn{2}{c}{\textbf{}}                         & \multicolumn{3}{c}{\textbf{After enforcing policy}} \\ \cmidrule{3-5} 
\textbf{Network Measures} & \multicolumn{1}{c}{\textbf{Initial}} & \textbf{China+1} & \textbf{Friendshoring} & \textbf{Reshoring} \\ \midrule
Avg. No. Edges                            & 56339     & 62015.8       & 67732            & 68218            \\
Density                                   & 0.000243  & 0.000268      & 0.000293         & 0.000295         \\
Degree Assortativity                      & -0.121478 & -0.124529     & -0.112532        & -0.105941        \\
Location Assortativity                    & 0.428589  & 0.390959      & 0.37841          & 0.633812         \\
Avg. Shortest Path Length                 & 4.575134  & 4.662426      & 4.621184         & 4.933708         \\
Avg. Domestic Connections                 & 1.836291  & 1.86969       & 2.036909         & 3.055271         \\
Avg. International Connections            & 1.866325  & 2.206007      & 2.414458         & 1.428036         \\
Clustering Coefficient                    & 0.025749  & 0.026326      & 0.02671          & 0.033874         \\
Modularity                                & 0.569907  & 0.521519      & 0.491539         & 0.661074         \\
No. Communities                           & 1446      & 1147.8        & 1427.6           & 1454.4           \\
No. Non-Substitutable Products (Mines \%) & 0         & 12 (0\%)      & 307 (8.47\%)     & 780 (17.44\%)    \\
No. Non-Substitutable Companies           & 0         & 128           & 634              & 1558             \\ \bottomrule
\end{tabular}%
}
\end{table}

\subsubsection{Geopolitical Scope}
\label{subsec:appendix_result_geopolitical}

\begin{table}[h]
\centering
\caption{Network measures before and after enforcing the three policies in the Low Risk cluster.}
\label{Tab:Tier1}
\resizebox{\textwidth}{!}{%
\begin{tabular}{llccc}
\toprule
\multicolumn{2}{l}{\textbf{}}                                 & \multicolumn{3}{c}{\textbf{After enforcing policy}}              \\ \cmidrule{3-5} 
\textbf{Network Measures}                 & \textbf{Initial} & \textbf{High Risk+1} & \textbf{Friendshoring} & \textbf{Reshoring} \\ \midrule
Avg. No. Edges                  & 56339     & 68467.2   & 68161     & 69999     \\
Density                         & 0.000243  & 0.000296  & 0.000294  & 0.000302  \\
Degree Assortativity            & -0.121478 & -0.13027  & -0.103142 & -0.093862 \\
Location Assortativity          & 0.428589  & 0.356546  & 0.372094  & 0.699809  \\
Avg. Shortest Path Length       & 4.575134  & 4.569131  & 4.729862  & 4.957766  \\
Avg. Domestic Connections       & 1.836291  & 1.915142  & 1.984569  & 3.381572  \\
Avg. International Connections  & 1.866325  & 2.584543  & 2.494992  & 1.218783  \\
Clustering Coefficient          & 0.025749  & 0.029307  & 0.027382  & 0.03714   \\
Modularity                      & 0.563223  & 0.484677  & 0.483396  & 0.677027  \\
No. Communities                 & 1438      & 1029.6    & 1408.8    & 1438      \\
No. Non-Substitutable Products (Mines \%) & 0         & 21 (0\%)   & 367 (9.54\%) & 831 (16.49\%) \\
No. Non-Substitutable Companies & 0         & 157       & 703       & 2243      \\ \bottomrule
\end{tabular}%
}
\end{table}

\begin{table}[h]
\centering
\caption{Network measures before and after enforcing the three policies in the Low \& Medium Risk cluster.}
\label{Tab:Tier12}
\resizebox{\textwidth}{!}{%
\begin{tabular}{lcccc}
\toprule
\multicolumn{2}{l}{\textbf{}}                                 & \multicolumn{3}{c}{\textbf{After enforcing policy}}              \\ \cmidrule{3-5} 
\textbf{Network Measures}                 & \textbf{Initial} & \textbf{High Risk+1} & \textbf{Friendshoring} & \textbf{Reshoring} \\ \midrule
Avg. No. Edges                  & 56339     & 72863.8   & 64989     & 72665     \\
Density                         & 0.000243  & 0.000315  & 0.000281  & 0.000314  \\
Degree Assortativity            & -0.121478 & -0.122476 & -0.109138 & -0.078241 \\
Location Assortativity          & 0.428589  & 0.334296  & 0.380879  & 0.752534  \\
Avg. Shortest Path Length       & 4.575134  & 4.563405  & 4.742278  & 5.093038  \\
Avg. Domestic Connections       & 1.836291  & 1.91873   & 1.892574  & 3.721215  \\
Avg. International Connections  & 1.866325  & 2.8699    & 2.378523  & 1.054351  \\
Clustering Coefficient          & 0.025749  & 0.029051  & 0.026587  & 0.040796  \\
Modularity                      & 0.566948  & 0.458832  & 0.498373  & 0.700344  \\
No. Communities                 & 1438      & 939.8     & 1248.8    & 1387.2    \\
No. Non-Substitutable Products (Mines \%) & 0         & 22 (0\%)   & 256 (8.59\%) & 840 (16.31\%) \\
No. Non-Substitutable Companies & 0         & 204       & 529       & 3508      \\ \bottomrule
\end{tabular}%
}
\end{table}

\begin{table}[h]
\centering
\caption{Top 5 most vulnerable products under each policy-scope scenario.}
\label{Tab:VulnerableProducts}
\resizebox{\textwidth}{!}{%
\begin{tabular}{lllll}
\toprule
\multicolumn{1}{c}{\textbf{Scope / Policy}} & \textbf{China+1} & \textbf{Friendshoring} & \textbf{Reshoring} \\ \midrule

\multirow{5}{*}{American} 
& Rear Axle Module      & Transmission    & Alumina \\
& ABS ECU              & Rebar                           & Transmission \\
& Brake Pedal          & Brake          & Rebar \\
& Air Bag Fabric       & Steel billets                   & Software \\
& Engine Cooling Module & Wheel Hub       & Wire rods \\ \midrule

\multirow{5}{*}{European} 
& Engine Cooling Module       & Dolomite                        & Steel meshes \\
& Engine Mount Rubber         & Automatic Transmission  & Steel \\
& Rear Axle Module           & Lead                            & Software\\
& Brake Pedal                 & Heat treatment &  Sinter\\
& Vehicle Dynamic Control System & Semi-soft coking coal &  Steel plates\\ \midrule

\multirow{5}{*}{Asian} 
& ABS ECU            & Rebar                          & Flat bars \\
& Brake Pedal        & Hydrogen tank                           & Tin \\
& Exhaust Module     & Assembly    & Tin plates\\
& Air Bag Fabric     & Refined lead     &  Remelt\\
& Rear Axle Module   &  Hot-rolled strip        &  Rebar\\ \midrule

\multicolumn{1}{c}{\textbf{Scope / Policy}} & \textbf{High Risk+1} & \textbf{Friendshoring} & \textbf{Reshoring} \\ \midrule

\multirow{5}{*}{Low Risk} 
& Engine Cooling Module& Rebar                          & Steel meshes \\
& Rear Axle Module     &  Sodium tripolyphosphate                          & Steel \\
& ABS ECU            & Wheel Hub         & Software \\
& Brake Pedal         &  Brake     & Alumina \\
& Air Bag Fabric      & Intake Manifold          & Remelt \\ \midrule

\multirow{5}{*}{Low \& Medium Risk} 
& Brake Pedal          & Rebar                           & Steel meshes \\
& Rear Axle Module     & Electrolyte for Lith. Ion Battery & Sinter \\
& ABS ECU             & Wheel Hub       & Software \\
& Engine Cooling Module& Zinc ingots                     & Steel \\
& Air Bag Fabric      & Brake Caliper  & Remelt \\

\bottomrule
\end{tabular}%
}
\end{table}

\end{document}